# On the Complexity of Decision Making in Possibilistic Decision Trees

**Hélène Fargier**
IRIT
118 route de Narbonne
31062 Toulouse, France
fargier@irit.fr

**Nahla Ben Amor**
LARODEC Laboratory
ISG, University of Tunis
Tunisia, 2000
nahla.benamor@gmx.fr

**Wided Guezguez**
LARODEC Laboratory
ISG, University of Tunis
Tunisia, 2000
widedguezguez@gmail.com

**Abstract**

When the information about uncertainty cannot be quantified in a simple, probabilistic way, the topic of possibilistic decision theory is often a natural one to consider. The development of possibilistic decision theory has lead to a series of possibilistic criteria, e.g pessimistic possibilistic qualitative utility, possibilistic likely dominance , binary possibilistic utility and possibilistic Choquet integrals. This paper focuses on sequential decision making in possibilistic decision trees. It proposes a complexity study of the problem of finding an optimal strategy depending on the monotonicity property of the optimization criteria which allows the application of dynamic programming that offers a polytime reduction of the decision problem. It also shows that possibilistic Choquet integrals do not satisfy this property, and that in this case the optimization problem is $NP-hard$.

## 1 Introduction

For several years, there has been a growing interest in the Artificial Intelligence community towards the foundations and computational methods of decision making under uncertainty. This is especially relevant for applications to sequential decision making under uncertainty, where a suitable strategy is to be found, that associate a decision to each state of the world. Several formalisms can be used for sequential decision problems, such as decision trees, influence diagrams or Markov decision processes. A decision tree is an explicit representation of a sequential decision problem, while influence diagrams or Markov decision processes are compact representations. Even in the simple and explicit case of decision trees, the set of potential strategies increases exponentially with the tree size.

A popular criterion to compare decisions under risk is the expected utility ($EU$) model axiomatized by von Neumann and Morgenstern [12]. This model relies on a probabilistic representation of uncertainty: an elementary decision (i.e. a one-step decision problem) is modeled by a probabilistic lottery over its possible outcomes. The preferences of the decision maker are supposed to be captured by a utility function assigning a numerical value to each outcome. The evaluation of a lottery is then performed through the computation of its expected utility (the greater is the better). In sequential decision making, each possible strategy is viewed as compound lottery. It can be reduced to an equivalent simple lottery, and thus compared to the others according to its expected utility. Although the high combinatorial nature of the set of possible strategies, the selection of an optimal strategy can be performed in polynomial time (polytime) with the size of the decision tree: the EU model indeed satisfies a property of *monotonicity* that guarantees completeness of a polytime algorithm of dynamic programming.

When the information about uncertainty cannot be quantified in a simple, probabilistic way the topic of possibilistic decision theory is often a natural one to consider [1, 2, 4]. Giving up the probabilistic quantification of uncertainty has led to give up the EU criterion as well. The development of possibilistic decision theory has lead a series of possibilistic counterparts of the EU criterion. [15] for instance advocates the use of possibilistic Choquet integrals, which relies on a numerical interpretation of both possibility and utility degrees. On the contrary, [4] have studied the case of a qualitative interpretation and propose two criteria based on possibility theory, an optimistic and a pessimistic one (denoted by $U_{opt}$ and $U_{pes}$), whose definitions only require a finite ordinal, non compensatory, scale for evaluating both utility and plausibility.

The axiomatization of $U_{opt}$ and $U_{pes}$ yielded the development of sophisticated qualitative models for sequential decision making, e.g. possibilistic Markov decision processes [16, 17], possibilistic ordinal decision trees [5] and even possibilistic ordinal influence diagrams. One of the most interesting properties of this qualitative model is indeed that it obeys a weak form of the monotonicity property. As a consequence, dynamic programming may be used and a strategy optimal with respect to $U_{pes}$ or $U_{opt}$ can be built in polytime.

On the contrary general Choquet integrals are incompatible with dynamic programming. Worst, the problem of determining a strategy optimal with respect to Choquet integrals is NP-Hard in the general case [10]. We show in the present paper that the problem of determining a strategy optimal with respect to a *possibilistic Choquet integrals* is NP-Hard as well. More generally, we propose a study of the complexity of the problem of finding an optimal strategy for possibilistic decision trees: which criteria obey the monotonicity property (and then may be solved in polytime thanks to dynamic programming) and which ones lead to NP-Hard problems?

This paper is organized as follows: Section 2 presents a refresher on possibilistic decision making under uncertainty and especially a survey on most common possibilistic decision criteria. Section 3 details possibilistic decision trees. Section 4 is devoted to the complexity study regarding different decision criteria. Finally, Section 5 presents an extension to order of magnitude expected utility. Proofs are omitted for space reasons but are available at `ftp://ftp.irit.fr/IRIT/ADRIA/PapersFargier/uai11.pdf`.

## 2 Background

### 2.1 Possibilitic lotteries

The basic building block in possibility theory is the notion of *possibility distribution* [3]. Let $x_1, \ldots, x_n$ be a set of state variables whose value are ill-known, $D_1 \ldots D_n$ their respective domains and denote $\Omega = D_1 \times \cdots \times D_n$ the joint domain of $x_1, \ldots, x_n$. Vectors $\omega \in \Omega$ are often called realizations or simply "states" (of the world). The agent's knowledge about the value of $x_i$'s is by a possibility distribution $\pi : \Omega \to [0, 1]$; $\pi(\omega) = 1$ means that $\omega$ is totally possible and $\pi(\omega) = 0$ means that $\omega$ is an impossible state. It is generally assumed that there exist at least one state $\omega$ which is totally possible, i.e. that $\pi$ is *normalized*.

Extreme cases of knowledge are presented by *complete knowledge* i.e. $\exists \omega_0$ s.t. $\pi(\omega_0) = 1$ and $\forall \omega \neq \omega_0, \pi(\omega) = 0$ and *total ignorance* i.e. $\forall \omega \in \Omega, \pi(\omega) = 1$ (all values in $\Omega$ are possible). From $\pi$, one can compute the possibility $\Pi(A)$ and necessity $N(A)$ of an event $A \subseteq \Omega$:

$$\Pi(A) = \sup_{\omega \in A} \pi(\omega). \qquad (1)$$

$$N(A) = 1 - \Pi(\bar{A}) = 1 - \sup_{\omega \notin A} \pi(\omega). \qquad (2)$$

$\Pi(A)$ evaluates to which extend $A$ is consistent with the knowledge represented by $\pi$; $N(A)$ corresponds to the extent to which $\neg A$ is impossible: it evaluates at which level $A$ is implied by the knowledge.

In possibility theory, conditioning is defined by the following counterpart of the Bayesian rule:

$$\forall \omega, \ \pi(\omega) = \min(\pi(\omega \mid \psi), \Pi(\psi)). \qquad (3)$$

In this equation, $\pi(\omega \mid \psi)$ and $\Pi(\psi)$ are combined according to a min operation, according to the *ordinal* interpretation of the possibilistic scale[1]. The following min-based definition of the conditioning corresponds to the least specific solution of (3) (see [9]):

$$\pi(\omega|\psi) = \begin{cases} 1 & if \ \pi(\omega) = \Pi(\psi) \ and \ \omega \in \psi \\ \pi(\omega) & if \ \pi(\omega) < \Pi(\psi) \ and \ \omega \in \psi \\ 0 & otherwise \end{cases} \qquad (4)$$

Following [2, 4], a decision can be seen as a possibility distribution over its outcomes. In a single stage problem, a utility function maps each outcome to a utility value in a totally ordered set $U = \{u_1, \ldots, u_n\}$ (we assume without loss of generality that $u_1 \leq \cdots \leq u_n$). This function models the attractiveness of each outcome for the decision maker. An act can then be represented by a possibility distribution on $U$, called a (simple) *possibilistic lottery*, and denoted by $\langle \lambda_1/u_1, \ldots, \lambda_n/u_n \rangle$: $\lambda_i = \pi(u_i)$ is the possibility that the decision leads to an outcome of utility $u_i$.

In the following, $\mathcal{L}$ denotes the set of simple lotteries (i.e. the set of possibility distributions over $U$). A *possibilistic compound lottery* $\langle \lambda_1/L_1, \ldots, \lambda_m/L_m \rangle$ (also denoted by $(\lambda_1 \wedge L_1 \vee \cdots \vee \lambda_m \wedge L_m)$) is a possibility distribution over a subset of $\mathcal{L}$. The possibility $\pi_{i,j}$ of getting a utility degree $u_j \in U$ from one of its sub−lotteries $L_i$ depends on the possibility $\lambda_i$ of getting $L_i$ and on the conditional possibility $\lambda_j^i = \pi(u_j \mid L_i)$ of getting $u_j$ from $L_i$ i.e. $\pi_{i,j} = \min(\lambda_j, \lambda_j^i)$ by equation (3). Hence, the possibility of getting $u_j$ from a compound lottery $\langle \lambda_1/L_1, \ldots, \lambda_m/L_m \rangle$ is the *max*, over all $L_i$, of $\pi_{i,j}$. Thus, [2, 4] have proposed to reduce $(\lambda_1/L_1, \ldots, \lambda_m/L_m)$ into a simple lottery, denoted by $Reduction(\langle \lambda_1/L_1, \ldots, \lambda_m/L_m \rangle)$, that is considered as equivalent to the compound one:

$$Reduction(\langle \lambda_1/L_1, \ldots, \lambda_m/L_m \rangle)$$
$$= \langle \max_{j=1..m} \min(\lambda_j, \lambda_1^j)/u_1, \ldots, \max_{j=1..m} min(\lambda_j, \lambda_n^j)/u_n \rangle. \qquad (5)$$

---

[1] The other, *numerical* interpretation of possibility theory is the use a product instead of a min operation, but this is out the scope of the present study.

Obviously, the reduction of a simple lottery is the simple lottery itself. Since min and max are polynomial operations, the reduction of a compound lottery is polynomial in the size of the compound lottery [2]. We review in the following the different criteria that have been proposed to evaluate and/or compare (simple) lotteries; thanks to the notion of reduction, they also apply to compound lotteries: to evaluate/compare compound lotteries, simply reduce each to an equivalent simple one; then use one of the criteria proposed for the evaluation/the comparison of simple lotteries. Formally, any comparison criterion $O$, i.e. any preference order $\geq_O$ defined on simple lotteries can be extended to compound lotteries as follows:

$$L \geq_O L' \iff Reduction(L) \geq_O Reduction(L'). \quad (6)$$

## 2.2 Possibilistic qualitative utilities ($U_{pes}, U_{opt}, PU$)

Under the assumption that the utility scale and the possibility scale are commensurate and purely ordinal, [4] have proposed the following qualitative pessimistic and optimistic utility degrees for evaluating any simple lottery $L = \langle \lambda_1/u_1, \ldots, \lambda_n/u_n \rangle$ (possibly issued from the reduction of a compound lottery):

$$U_{pes}(L) = \max_{i=1..n} \min(u_i, N(L \geq u_i)). \quad (7)$$

$$U_{opt}(L) = \max_{i=1..n} \min(u_i, \Pi(L \geq u_i)). \quad (8)$$

where $N(L \geq u_i) = 1 - \Pi(L < u_i) = 1 - \max_{j=1,i-1} \lambda_j$ and $\Pi(L \geq u_i) = \max_{j=1..n} \lambda_j$ are the necessity and possibility degree that $L$ reaches at least the utility value $u_i$. $U_{pes}$ generalizes the *Wald criterion* and estimates to what extend it is certain (i.e. necessary according to measure $N$) that $L$ reaches a good utility. Its optimistic counterpart, $U_{opt}$, estimates to what extend it is possible that $L$ reaches a good utility. Because decision makers are rather cautious than adventurous, the former is generally preferred to the latter.

Claiming that the lotteries that realize in the best prize or in the worst prize play an important role in decision making, Giang and Shenoy [7] have proposed a bipolar model in which the utility of an outcome is a pair $u = \langle \overline{u}, \underline{u} \rangle$ where $max(\overline{u}, \underline{u}) = 1$: the utility is binary in this sense that $\overline{u}$ is interpreted as the possibility of getting the ideal, good reward (denoted $\top$) and $\underline{u}$ is interpreted as the possibility of getting the anti ideal, bad reward (denoted $\bot$).

Because of the normalization constraint $max(\overline{u}, \underline{u}) = 1$, the set $U = \{\langle \overline{u}, \underline{u} \rangle \in [0,1]^2, max(\lambda, \mu) = 1\}$ is totally ordered:

$$\langle \overline{u}, \underline{u} \rangle \succeq_b \langle \overline{v}, \underline{v} \rangle \iff \begin{cases} \overline{u} = \overline{v} = 1 \text{ and } \underline{u} \leq \underline{v} \\ \text{or} \\ \overline{u} \geq \overline{v} \text{ and } \underline{u} = \underline{v} = 1 \\ \text{or} \\ \overline{u} = \underline{v} = 1 \text{ and } \overline{v} < 1 \end{cases} \quad (9)$$

Each $u_i = \langle \overline{u_i}, \underline{u_i} \rangle$ in the utility scale is thus understood as a small lottery $\langle \overline{u_i}/\top, \underline{u_i}/\bot \rangle$. A lottery $\langle \lambda_1/u_1, \ldots, \lambda_n/u_n \rangle$ can be viewed as a compound lottery, and its *PU* utility is computed by reduction:

$$PU(\langle \lambda_1/u_1, \ldots, \lambda_n/u_n \rangle)$$
$$= Reduction(\lambda_1/\langle \overline{u_1}/\top, \underline{u_1}/\bot \rangle, \ldots, \lambda_n \wedge \langle \overline{u_n}/\top, \underline{u_n}/\bot \rangle)$$
$$= \langle \max_{j=1..n}(\min(\lambda_j, \overline{u_j}))/\top, \max_{j=1..n}(\min(\lambda_j, \underline{u_j}))/\bot \rangle \quad (10)$$

We thus get, for any lottery $L$ a binary utility $PU(L) = \langle \overline{u}, \underline{u} \rangle$ in $U$. Lotteries can then be compared according to Equation (9):

$$L \geq_{PU} L' \iff Reduction(L) \succeq_b Reduction(L'). \quad (11)$$

In [8] Giang and Shenoy show that the order induced by PU collapse with the one induced by $U_{opt}$ whenever for any lottery, the possibility $\underline{u}$ of getting the worst utility is equal to 1 (any "compound" lottery $\lambda_1/\langle 0, \alpha_1 \rangle, \ldots, \lambda_n/\langle 0, \alpha_n \rangle$ reduces to the lottery $\langle 1, \max_{i=1..n} min(\lambda_i, \alpha_i) \rangle$ : $\max_{i=1..n} min(\lambda_i, \alpha_i)$ is precisely the optimistic utility value). In the same way, Giang and Shenoy have shown that the order induced on the lotteries by PU collapse with the one induced by $U_{pes}$ as soon as for any lottery, the possibility $\overline{u}$ of getting the best utility is equal to 1. We shall thus say that PU captures $U_{opt}$ and $U_{pes}$ as particular cases.

## 2.3 Possibilistic likely dominance ($LN, L\Pi$)

When the scales evaluating the utility and the possibility of the outcomes are not commensurate, [1, 6] propose to prefer, among two possibilistic decisions, the one that is more likely to overtake the other. Such a rule does not assign a global utility degree to the decisions, but draws a pairwise comparison. Although designed on a Savage-like framework rather than on lotteries, it can be translated on lotteries. This rule states that given two lotteries $L_1 = \langle \lambda_1^1/u_1^1, \ldots, \lambda_n^1/u_n^1 \rangle$ and $L_2 = \langle \lambda_1^2/u_1^2, \ldots, \lambda_n^2/u_n^2 \rangle$, $L_1$ is as least as good as $L_2$ as soon as the likelihood (here, the necessity or the possibility) of the event *the utility of $L_1$ is as least as good as the utility of $L_2$* is greater or equal to the likelihood of the event *the utility of $L_2$ is as least as good as the utility of $L_1$*. Formally:

$$L_1 \geq_{LN} L_2 \iff N(L_1 \geq L_2) \geq N(L_2 \geq L_1), \quad (12)$$

---

[2] The size of a simple lottery is the number of its outcomes; the size of a compound lottery is the sum of the sizes of its sub-lotteries plus the number of its sub-lotteries.

$$L_1 \geq_{L\Pi} L_2 \iff \Pi(L_1 \geq L_2) \geq \Pi(L_2 \geq L_1) \quad (13)$$

where $\Pi(L_1 \geq L_2) = \sup_{u_i^1, u_i^2 \text{ s.t. } u_i^1 \geq u_i^2} min(\lambda_i^1, \lambda_i^2)$ and $N(L_1 \geq L_2) = 1 - \sup_{u_i^1, u_i^2 \text{ s.t. } u_i^1 < u_i^2} min(\lambda_i^1, \lambda_i^2)$.

The preference order induced on the lotteries is not transitive, but only quasitransitive: obviously $L_1 >_N L_2$ and $L_2 >_{LN} L_3$ implies $L_1 >_{LN} L_3$ (resp. $L_1 >_{L\Pi} L_2$ and $L_2 >_{L\Pi} L_3$ implies $L_1 >_{L\Pi} L_3$) but it may happen that $L_1 \sim_{LN} L_2$, $L_2 \sim_{LN} L_3$ (resp. $L_1 \sim_{L\Pi} L_2$, $L_2 \sim_{L\Pi} L_3$) and $L_1 >_{LN} L_3$ (resp. $L_1 >_{L\Pi} L_3$).

### 2.4 Possibilistic Choquet integrals

In presence of heterogeneous information, i.e. when the knowledge about the state of the world is possibilistic while the utility degrees are numerical and compensatory the previous models cannot be applied anymore. Following [18] Choquet integrals appear as a right way to extend expected utility to non Bayesian models. Like the EU model, this model is a numerical, compensatory, way of aggregating uncertain utilities. But it does not necessarily resort on a Bayesian modeling of uncertain knowledge. Indeed, this approach allows the use of any monotonic set function [3], and thus of a necessity measure.

As the qualitative case, but assuming that the utility degrees have a richer, cardinal interpretation, the utility of $L$ is given by a Choquet integrals:

$$Ch_\mu(L) = \Sigma_{i=1,n}(u_i - u_{i-1}) \cdot \mu(L \geq u_i). \quad (14)$$

If $\mu$ is a probability measure then $Ch_\mu(L)$ is simply the expected utility of $L$. In the present paper, we are interested in studying the possibilistic framework for decision making: for cautious (resp. adventurous) decision makers, the capacity $\mu$ is the necessity measure $N$ (resp. the possibility measure $\Pi$):

$$Ch_N(L) = \Sigma_{i=1,n}(u_i - u_{i-1}) \cdot N(L \geq u_i). \quad (15)$$

$$Ch_\Pi(L) = \Sigma_{i=1,n}(u_i - u_{i-1}) \cdot \Pi(L \geq u_i). \quad (16)$$

From Equations (5) and (15), it can be shown that:

**Proposition 1.**
Given a lottery $L = \langle \lambda_1/u_1, \ldots, \lambda_n/u_n \rangle$, an utility $u_i$ s.t. $u_i \leq \max_{u_j \in L, \lambda_j > 0} u_j$ and a lottery $L' = \langle \lambda_1'/u_1, \ldots, \lambda_n'/u_n \rangle$ s.t. $\lambda_i' \geq \lambda_i$ and $\forall j \neq i, \lambda_j' = \lambda_j$, it holds that $Ch_N(L') \leq Ch_N(L)$.

This emphasizes the pessimistic character of $Ch_N$: adding to a lottery any consequence that is not better than its best one decreases its evaluation. As a consequence, we get the following result

**Proposition 2.** Let $L_1$, $L_2$ be two lotteries such that $\max_{u_i \in L_2, \lambda_i > 0} u_i \leq \max_{u_i \in L_1, \lambda_i > 0} u_i$. It holds that $Ch_N(Reduction(\langle 1/L_1, 1/L_2 \rangle)) \leq Ch_N(L_1)$.

No such property holds for $Ch_\Pi$, as shown by the following counter example:

**Counter Example 1.** Let $U = \{0, 1, \ldots, 9\}$, $L_1 = \langle 0.2/0, 1/2, 0.5/9 \rangle$, $L_2 = \langle 0.4/4, 1/7 \rangle$[4] and $L_3 = Reduction(\langle 1/L_1, 1/L_2 \rangle) = \langle 0.2/0, 1/2, 0.4/4, 1/7, 0.5/9 \rangle$. We can check that $Ch_\Pi(L_1) = 5.5$ and $Ch_\Pi(L_3) = 8$.

## 3 Possibilistic decision trees

Decision trees are graphical representations of sequential decision problems under the assumption of full observability. A decision tree is a tree $\mathcal{T} = (\mathcal{N}, \mathcal{E})$ whose set of nodes, $\mathcal{N}$, contains three kinds of nodes:

- $\mathcal{D} = \{D_0, \ldots, D_m\}$ is the set of decision nodes (represented by rectangles). The labeling of the nodes is supposed to be in accordance with the temporal order i.e. if $D_i$ is a descendant of $D_j$, then $i > j$. The root node of the tree is necessarily a decision node, denoted by $D_0$.

- $\mathcal{LN} = \{LN_1, \ldots, LN_k\}$ is the set of leaves, also called utility leaves: $\forall LN_i \in \mathcal{LN}$, $u(LN_i)$ is the utility of being eventually in node $LN_i$. For the sake of simplicity we assume that only leave nodes lead to utilities.

- $\mathcal{C} = \{C_1, \ldots, C_n\}$ is the set of chance nodes represented by circles. For any $X_i \in \mathcal{N}$, let $Succ(X_i) \subseteq \mathcal{N}$ be the set of its children. For any $D_i \in \mathcal{D}, Succ(D_i) \subseteq \mathcal{C}$: $Succ(D_i)$ is the set of actions that can be decided when $D_i$ is observed. For any $C_i \in \mathcal{C}, Succ(C_i) \subseteq \mathcal{LN} \cup \mathcal{D}$: $Succ(C_i)$ is the set of outcomes of the action $C_i$ - either a leaf node is observed, or a decision node is observed (and then a new action should be executed).

The size of $\mathcal{T}$ is its number of edges (the number of nodes is equal to the number of edges plus 1).

In classical, probabilistic, decision trees the uncertainty pertaining to the possible outcomes of each $C_i \in \mathcal{C}$, is represented by a conditional probability distribution $p_i$ on $Succ(C_i)$, such that $\forall N \in Succ(C_i)$, $p_i(N) = P(N|path(C_i))$ where $path(C_i)$ denotes all the value assignments to chance and decision nodes on the path from the root node to $C_i$. In this work,

---
[3]This kind of set function is often called capacity or fuzzy measure.

[4]For the sake of simplicity, we shall forget about the utility degrees that receive a possibility degree equal to 0 in a lottery, i.e. we write $\langle 0.2/0, 1/2, 0.5/9 \rangle$ instead of $\langle 0.2/0, 1/2, 0/3, 0/4, 0/5, 0/6, 0/7, 0/8, 0.5/9 \rangle$.

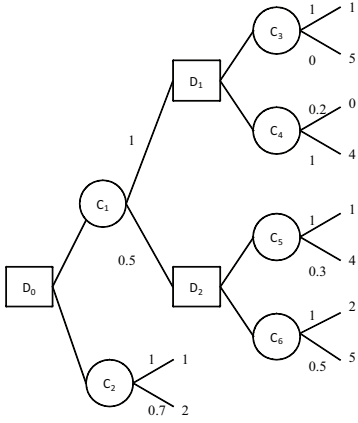

Figure 1: Example of possibilistic decision tree with $\mathcal{C} = \{C_1, C_2, C_3, C_4, C_5, C_6\}$, $\mathcal{D} = \{D_0, D_1, D_2\}$ and $\mathcal{LN} = U = \{0, 1, 2, 3, 4, 5\}$.

we obviously use a possibilistic labeling (see Figure 1). The difference with probabilistic decision trees is that the chance nodes are viewed as possibilistic lotteries. More precisely, for any $C_i \in \mathcal{C}$, the uncertainty pertaining to the more or less possible outcomes of each $C_i$ is represented by a *conditional possibility distribution* $\pi_i$ on $Succ(C_i)$, such that $\forall N \in Succ(C_i), \pi_i(N) = \Pi(N|path(C_i))$.

Solving the decision tree amounts at building a *strategy* that selects an action (i.e. a chance node) for each reachable decision node. Formally, we define a strategy as a function $\delta$ from $\mathcal{D}$ to $\mathcal{C} \cup \{\bot\}$. $\delta(D_i)$ is the action to be executed when a decision node $D_i$ is observed. $\delta(D_i) = \bot$ means that no action has been selected for $D_i$ (because either $D_i$ cannot be reached or the strategy is partially defined). Admissible strategies must be:
- *sound*: $\forall D_i \in \mathcal{D}, \delta(D_i) \in Succ(D_i) \cup \{\bot\}$.
- *complete*: (i) $\delta(D_0) \neq \bot$ and (ii) $\forall D_i$ s.t. $\delta(D_i) \neq \bot, \forall N \in Succ(\delta(D_i))$, either $\delta(N) \neq \bot$ or $N \in \mathcal{LN}$.

Let $\Delta$ be the set of sound and complete strategies that can be built from $\mathcal{T}$. Any strategy in $\Delta$ can be viewed as a connected subtree of $\mathcal{T}$ whose edges are of the form $(D_i, \delta(D_i))$. The size of a strategy $\delta$ is the sum of its number of nodes and edges, it is obviously lower than the size of the decision tree.

Strategies can be evaluated and compared thanks to the notion of lottery reduction. Recall indeed that leaf nodes in $\mathcal{LN}$ are labeled with utility degrees. Then a chance node can be seen as a simple lottery (for the most right chance nodes) or as a compound lottery (for the inner chance nodes). Each strategy is a compound lottery and can be reduced to an equivalent simple one. Formally, the composition of lotteries will be applied from the leafs of the strategy to its root, according to the following recursive definition for any $N_i$ in $\mathcal{N}$:

$$L(N_i, \delta) = \begin{cases} L(\delta(N_i), \delta) & if \ N_i \in \mathcal{D} \\ Reduction(\langle \pi_i(X_j)/L(X_j, \delta)_{X_j \in Succ(N_i)} \rangle) \\ \quad if \ N_i \in \mathcal{C} \\ \langle 1/u(N_i) \rangle & if \ N_i \in \mathcal{LN} \end{cases} \quad (17)$$

Equation (17) is simply the adaptation to strategies of lottery reduction (Equation (5)). We can then compute $Reduction(\delta) = L(D_0, \delta)$: $Reduction(\delta)(u_i)$ is simply the possibility of getting utility $u_i$ when $\delta$ is applied from $D_0$. Since, the operators max and min are polytime Equation (17) define a polytime computation of the reduced lottery.

**Proposition 3.** *For any strategy $\delta$ in $\Delta$, an equivalent simple possibilistic lottery can be computed in polytime.*

We are now in position to compare strategies, and thus to define the notion of optimality. Let $O$ be one of the criteria defined in Section 2 (i.e. depending on the application, $\geq_O$ is either $\geq_{L\Pi}$, or $\geq_{LN}$, or the order induced by $U_{pes}$, or by $U_{opt}$, etc.). A strategy $\delta \in \Delta$, is said to be optimal w.r.t. $\geq_O$ iff:

$$\forall \delta' \in \Delta, Reduction(\delta) \geq_O Reduction(\delta'). \quad (18)$$

Notice that this definition does not require the full transitivity (nor the completeness) of $\geq_O$ and is meaningful as soon as the strict part of $\geq_O$, $>_O$, is be transitive. This means that it is applicable to the preference relations that rely on the comparison of global utilities (qualitative utilities, binary utility, Choquet integrals) but also to $\geq_{LN}$ and $\geq_{L\Pi}$. We show in the following that the complexity of the problem of optimization depends on the criterion at work.

## 4 On the complexity of decision making in possibilistic decision trees

Finding optimal strategies via an exhaustive enumeration of $\Delta$ is a highly computational task. For instance, in a decision tree with $n$ decision nodes and a branching factor equal to 2, the number of potential strategies is in $O(2^{\sqrt{n}})$. For standard probabilistic decision trees, where the goal is to maximize expected utility, an optimal strategy can be computed in polytime thanks to an algorithm of dynamic programming which builds the best strategy backwards, optimizing the decisions from the leaves of the tree to its root.

Regarding possibilistic decision trees, [5] shows that such a method can also be used to get a strategy max-imizing $U_{pes}$ and $U_{opt}$. The reason is that like EU, $U_{pes}$ satisfies the key property of weak monotonicity. We formulate it for any criterion $O$ over possibilistic lotteries: $\geq_O$ is said to be weakly monotonic iff whatever $L, L'$ and $L"$, whatever $(\alpha,\beta)$ such that $\max(\alpha, \beta) = 1$:

$$L \geq_O L' \Rightarrow (\alpha \wedge L) \vee (\beta \wedge L") \geq_O (\alpha \wedge L') \vee (\beta \wedge L"). \quad (19)$$

This property states that the combination of $L$ (resp. $L'$) with $L"$, does not change the initial order induced by $O$ between $L$ and $L'$ - this allows dynamic programming to decide in favor of $L$ or $L'$ before considering the compound decision. The principle of backwards reasoning procedure is depicted in a recursive manner by Algorithm 1 for any preference order $\geq_O$ among lotteries. When each chance node is reached, an optimal sub-strategy is built for each of its children - these sub-strategies are combined w.r.t. their possibility degrees, and the resulting compound strategy is reduced: we get an equivalent simple lottery, representing the current optimal sub-strategy. When a decision node $X$ is reached, a decision $Y^*$ leading to a sub-strategy optimal w.r.t. $\geq_O$ is selected among all the possible decisions $Y \in Succ(X)$, by comparing the simple lotteries equivalent to each sub strategies.

This procedure crosses each edge in the tree only once. When the comparison of simple lotteries by $\geq_O$ (Line (2)) and the reduction operation on a 2-level lottery (Line (1)) can be performed in polytime, its complexity is polynomial w.r.t. the size of the tree. Then:

**Proposition 4.** *If $\geq_O$ satisfies the monotonicity property, then Algorithm 1 computes a strategy optimal w.r.t. $O$ in polytime.*

We will see in the following that, beyond $U_{pes}$ and $U_{opt}$ criteria, several other criteria satisfy the monotonicity property and that their optimization can be managed in polytime by dynamic programming. The possibilistic Choquet integrals, on the contrary, do not satisfy weak monotonicity; we will show that they lead to NP-Complete decision problems. Formally, for any of the optimization criteria of Sections 2.2 to 2.4, the corresponding decision problem can be defined as follows:

[DT-OPT-$O$] (Strategy optimization w.r.t. an optimization criterion $O$ in possibilistic decision trees)
INSTANCE: A possibilistic decision tree $\mathcal{T}$, a level $\alpha$.
QUESTION: Does there exist a strategy $\delta \in \Delta$ such as $Reduction(\delta) \geq_O \alpha$?

For instance DT-OPT-$Ch_N$ corresponds to the optimization of the necessity-based Choquet integrals. DT-OPT-$U_{pes}$ and DT-OPT-$U_{opt}$ correspond to the optimization of the possibilistic qualitative utilities $U_{pes}$ and $U_{opt}$, respectively.

**Algorithm 1:** Dynamic programming
Data: In: a node $X$, In/Out: a strategy $\delta$
Result: A lottery $L$
begin
    for $i \in \{1, \ldots, n\}$ do $L[u_i] \leftarrow 0$
    if $N \in \mathcal{LN}$ then $L[u(N)] \leftarrow 1$
    if $N \in \mathcal{C}$ then
        % Reduce the compound lottery
        foreach $Y \in Succ(N)$ do
            $L_Y \leftarrow ProgDyn(Y, \delta)$
            for $i \in \{1, \ldots, n\}$ do
                $L[u_i] \leftarrow$
                $\max(L[u_i], \min(\pi_N(Y), L_y[u_i]))$ (Line (1))
    if $N \in \mathcal{D}$ then
        % Choose the best decision
        $Y^* \leftarrow Succ(N).first$
        foreach $Y \in Succ(N)$ do
            $L^Y \leftarrow ProgDyn(Y, \delta)$
            if $L_Y >_O L_{Y^*}$ then $Y^* \leftarrow Y$ (Line (2))
        $\delta(N) \leftarrow Y^*$
        $L \leftarrow L_{Y^*}$
    return $L$
end

### 4.1 Possibilistic qualitative utilities ($U_{pes}, U_{opt}, PU$)

Possibilistic qualitative utilities $U_{pes}$ and $U_{opt}$ satisfy the weak monotonicity principle. Although not referring to a classical, real-valued utility scale, but to a 2 dimensional scale, this is also the case of $PU$.

**Proposition 5.** $\geq_{U_{pes}}$, $\geq_{U_{opt}}$ and $\geq_{PU}$ satisfy the weak monotonicity property.

As a consequence, dynamic programming applies to the optimization of these criteria in possibilistic decision trees. Although not explicitly proved in the literature, Proposition 5 is common knowledge in possibilistic decision theory. It is also known that dynamic programming applies to the optimization of $U_{pes}$, $U_{opt}$ and $PU$ in possibilistic Markov decision processes and thus to decision trees (see [5, 14, 17]).

**Corollary 1.** *DT-OPT-$U_{pes}$, DT-OPT-$U_{opt}$ and DT-OPT-PU belong to P.*

### 4.2 Possibilistic likely dominance ($LN, L\Pi$)

Fortunately, the optimization of the possibilistic likely dominance criteria also belongs to P. Indeed:

**Proposition 6.** $\geq_{L\Pi}$ and $\geq_{LN}$ satisfy the weak monotony principle

Algorithm 1 is thus sound and complete for $\geq_{L\Pi}$ and $\geq_{LN}$, and provides in polytime any possibilistic decision tree with a strategy optimal w.r.t. these criteria.

**Corollary 2.** $DT-OPT-LN$ and $DT-OPT-L\Pi$ belong to $P$.

It should be noticed that, contrarily to what can be done with the three previous rules, the likely dominance comparison of two lotteries will be reduced to a simple comparison of aggregated values (Line (2)) Anyway, since only one best strategy is looked for, the transitivity of $>_{L\Pi}$ (resp. $>_{L\Pi}$) guarantees the correctness of the procedure - the non transitivity on the indifference is not a handicap when only one among the best strategies is looked for. The difficulty would be raised if we were looking for all the best strategies.

### 4.3 Possibilistic Choquet integrals ($Ch_N$, $Ch_\Pi$)

The situation is thus very confortable with qualitative utilities, binary possibilistic utility and likely dominance. It is much lesser comfortable for the case of numerical utilities, i.e. when the aim is to optimize a possibilistic Choquet integral (either $Ch_N$ or $Ch_\Pi$). The point is that the possibilistic Choquet integrals (as many other Choquet integrals) do no satisfy the monotonicity principle:

**Counter Example 2** ( From [11] ). Let $L = \langle 0.2/0, 0.5/0.51, 1/1\rangle$, $L' = \langle 0.1/0, 0.6/0.5, 1/1\rangle$ and $L" = \langle 0.01/0, 1/1\rangle$.
$L_1 = (\alpha \wedge L) \vee (\beta \wedge L")$ and $L_2 = (\alpha \wedge L') \vee (\beta \wedge L")$, with $\alpha = 0.55$ and $\beta = 1$. Using Equation (5) we have: $Reduction(L_1) = \langle 0.2/0, 0.5/0.51, 1/1\rangle$ and $Reduction(L_2) = \langle 0.1/0, 0.55/0.5, 1/1\rangle$.
Computing $Ch_N(L) = 0.653$ and $Ch_N(L') = 0.650$ we get $L \geq_{Ch_N} L'$. But $Ch_N(Reduction(L_1)) = 0.653 < Ch_N(Reduction(L_3)) = 0.675$, i.e. $(\alpha \wedge L) \vee (\beta \wedge L") <_{Ch_N} (\alpha \wedge L') \vee (\beta \wedge L")$: this contradicts the monotonicity property.

Let $L = \langle 1/0, 0.5/0.51, 0.2/1\rangle$, $L' = \langle 1/0, 0.6/0.5, 0.1/1\rangle$ and $L" = \langle 1/0, 0.49/0.51\rangle$.
$L_1 = (\alpha \wedge L) \vee (\beta \wedge L")$ and $L_2 = (\alpha \wedge L') \vee (\beta \wedge L")$, with $\alpha = 1$ and $\beta = 0.55$. Using Equation (5) we have: $Reduction(L_1) = \langle 1/0, 0.5/0.51, 0.2/1\rangle$ and $Reduction(L_2) = \langle 1/0, 0.6/0.5, 0.49/0.51, 0.1/1\rangle$.
Computing $Ch_\Pi(L) = 0.353$ and $Ch_\Pi(L') = 0.350$ we get $L >_{Ch_\Pi} L'$. But $Ch_\Pi(Reduction(L_1)) = 0.3530 < Ch_\Pi(Reduction(L_2)) = 0.3539$, i.e. $(\alpha \wedge L) \vee (\beta \wedge L") <_{Ch_\Pi} (\alpha \wedge L') \vee (\beta \wedge L")$: this contradicts the monotonicity property.

**Proposition 7.** $DT-OPT-Ch_N$ and $DT-OPT-Ch_\Pi$ are NP-Complete.

## 5 Extension to Order of Magnitude Expected Utility

*Order of Magnitude Expected Utility* theory relies on a qualitative representation of beliefs, initially proposed by Spohn [19], via *Ordinal Conditional Functions*, and later popularized under the term *kappa-rankings*. $\kappa$ : $2^\Omega \to Z^+ \cup \{+\infty\}$ is a kappa-ranking if and only if:

S1 $\min_{\omega \in \Omega} \kappa(\{\omega\}) = 0$

S2 $\kappa(A) = \min_{\omega \in A} \kappa(\{\omega\})$ if $\emptyset \neq A \subseteq A$, $\kappa(\emptyset) = +\infty$

Note that event $A$ is more likely than event $B$ if and only if $\kappa(A) < \kappa(B)$: kappa-rankings have been termed as *disbelief functions*. They receive an interpretation in terms of order of magnitude of *small* probabilities. $\kappa(A) = i$ is equivalent to $P(A)$ is of the same order of $\varepsilon^i$, for a given fixed infinitesimal $\varepsilon$. There exists a close link between kappa-rankings and possibility measures, insofar as any kappa-ranking can be represented by a possibility measure, and vice versa.

Order of magnitude utilities have been defined in the same way [13, 20]. Namely, an order of magnitude function $\mu : X \to Z^+ \cup \{+\infty\}$ can be defined in order to rank outcomes $x \in X$ in terms of degrees of "dissatisfaction". Once again, $\mu(x) < \mu(x')$ if and only if $x$ is more desirable than $x'$, $\mu(x) = 0$ for the most desirable consequences, and $\mu(x) = +\infty$ for the least desirable consequences. $\mu$ is interpreted as: $\mu(x) = i$ is equivalent to say that the utility of $x$ is of the same order of $\varepsilon^i$, for a given fixed infinitesimal $\varepsilon$. An *order of magnitude expected utility* (OMEU) model can then be defined (see [13, 20] among others). Considering an order of magnitude lottery $L = \langle \kappa_1/\mu_1, \ldots, \kappa_n/\mu_n\rangle$ as representing a some probabilistic lottery, it is possible to compute the order of magnitude of the expected utility of this probabilistic lottery: it is equal to $\min_{i=1,n}\{\kappa_i + \mu_i\}$. Hence the definition of the OMEU value of a $\kappa$ lottery $L = \langle \kappa_1/\mu_1, \ldots, \kappa_n/\mu_n\rangle$:

$$OMEU(L) = \min_{i=1,n}\{\kappa_i + u_i\}. \quad (20)$$

The preference relation $\geq_{OMEU}$ is thus defined as:

$$L \geq_{OMEU} L' \iff OMEU(L) \leq OMEU(L'). \quad (21)$$

We shall now define kappa decision trees: for any $C_i \in \mathcal{C}$, the uncertainty pertaining to the more or less possible outcomes of each $C_i$ is represented by a kappa degree $\kappa_i(N) = Magnitude(P(N|past(C_i))), \forall N \in Succ(C_i)$ (with the normalization condition that the degree $\kappa = 0$ is given to at least one $N$ in $Succ(C_i)$). According to the interpretation of kappa ranking in terms of order of magnitude of probabilities, the product of infinitesimal the conditional probabilities along the paths lead to a sum of the kappa levels. Hence the following principle of reduction of the kappa lotteries:

$$Reduction(\kappa_1 \wedge L_1 \vee \cdots \vee \kappa_m \wedge L_m)$$
$$= \langle \min_{j=1..m}(\kappa_1^j + \kappa_j)/u_1, \ldots, \min_{j=1..m}(\kappa_n^j + \kappa_j)/u_n\rangle. \quad (22)$$

The last result of this paper is that OMEU satisfies the weak monotonicity principle:

**Proposition 8.** $\forall L, L', L" \in \mathcal{L}$,
$OMEU(L) \geq OMEU(L')$
$\Rightarrow OMEU((\alpha \wedge L) \vee (\beta \wedge L")) \geq OMEU((\alpha \wedge L') \vee (\beta \wedge L"))$

As a consequence dynamic programming is sound and complete for the optimization of Order of Magnitude Expected Utility:

**Corollary 3.** *There exists a polynomial algorithm for finding a strategy optimal w.r.t. the Order of Magnitude Expected Utility for any kappa decision tree.*

## 6 Conclusion

In this paper, we have shown that the strategy optimization in possibilistic decision trees is tractable for most of the criteria, extending the results about the qualitative utility criteria to other criteria, namely the likely dominance rule. We have also shown that the problem is intractable for the Choquet-based criteria. Finally, we have extended this work to OMEU, defining a new model for sequential decision trees, extending the notion of reduction to kappa lotteries and showing that this models obey the weak monotonicity principle. These results are summarized in Table 1. It should be noticed that the optimization of the possibilistic Choquet integrals is "only" NP-hard: the computation of the Choquet value of a possibilistic strategy is polynomial, whereas this computation can be more costly for other capacity measures; for instance computing the Choquet value of a strategy on the basis of its multi prior expected utility is itself a NP-hard problem - and the corresponding optimization problem is probably beyond NP. So, it appears that the use of possibilistic decision criteria does not lead to an increase in complexity, except for Choquet integrals. This is good news and allows the extension of our work to possibilistic decision diagrams. Concerning the Choquet case, further work includes the development of a direct evaluation algorithm for possibilistic influence diagrams where possibilistic Choquet integrals are used as a decision criteria inspired by the variable elimination approach.

Table 1: Results about the complexity of $DT - OPT$.

| $U_{pes}$ | $U_{opt}$ | $PU$ | $Ch_N$ | $Ch_\Pi$ | $L\Pi$ | $LN$ | OMEU |
|---|---|---|---|---|---|---|---|
| P | P | P | NP-hard | NP-hard | P | P | P |